\documentclass[10pt,twocolumn,letterpaper]{article}
\usepackage{cvpr} % uncomment this for the final submission
\usepackage{times}
\usepackage{epsfig}
\usepackage{graphicx}
\usepackage{amsmath}
\usepackage{amssymb}
\usepackage{multirow}
\usepackage{booktabs}
\usepackage{xcolor}
\usepackage{colortbl}
\usepackage{makecell}
\usepackage{balance}

\definecolor{lightblue}{RGB}{173,216,230}
\newcommand{\modelname}{DiffSwap++}

\begin{document}

%%%%%%%%% TITLE
\title{DiffSwap++: 3D Latent-Controlled Diffusion for \\
Identity-Preserving Face Swapping}

\author{
    Weston Bondurant \quad 
    Arkaprava Sinha \quad 
    Hieu Le \quad 
    Srijan Das \quad 
    Stephanie Schuckers\\
    The University of North Carolina at Charlotte \\
    {\tt\small wbondura@charlotte.edu}
}

\maketitle
\thispagestyle{empty}

%%%%%%%%% ABSTRACT
\begin{abstract}
   Diffusion-based face swapping has recently demonstrated superior visual quality compared to GAN-based approaches; however, existing methods often suffer from fine-grained artifacts and weak identity preservation due to limited modeling of 3D facial structure. To address these limitations, we propose \textbf{DiffSwap++}, a novel face-swapping pipeline that leverages \textbf{3D-aware latent representations} to guide the diffusion generation process, enabling improved disentanglement of identity from pose and expression. Our method conditions the diffusion architecture on \textbf{source identity embeddings} and \textbf{target facial landmarks}, producing high-fidelity swaps that preserve the source identity while maintaining target attributes. Extensive experiments on three public datasets, supported by biometric evaluations and human user studies, demonstrate that DiffSwap++ consistently outperforms prior state-of-the-art methods in identity retention and perceptual realism. Code will be made publicly available at https://github.com/WestonBond/DiffSwapPP
\end{abstract}\vspace{-0.3in}

%%%%%%%%% BODY TEXT
\section{Introduction}

\label{sec:intro}
Face swapping is the task of transferring the identity from a source face image onto a target face image while preserving the \textit{background}, \textit{pose}, and \textit{expression} of the target. Earlier methods have shown promising results, through the use of GAN-based architectures that integrate identity features from the source image with attribute features (e.g., \textit{pose}, \textit{expression}) from the target image \cite{simswap, hififace, faceshifter, hireslatent}. However, GAN training is inherently unstable, and qualitative evaluations reveal noticeable limitations in the realism and consistency of generated swapped faces\cite{spectral}. To address the training challenges and generation artifacts associated with GANs, recent approaches have explored the use of diffusion models for face swapping, demonstrating improved fidelity and robustness \cite{diffface, diffswap, reface}.

\begin{figure}
    \centering
    \scalebox{0.8}{
    \includegraphics[width=1\linewidth]{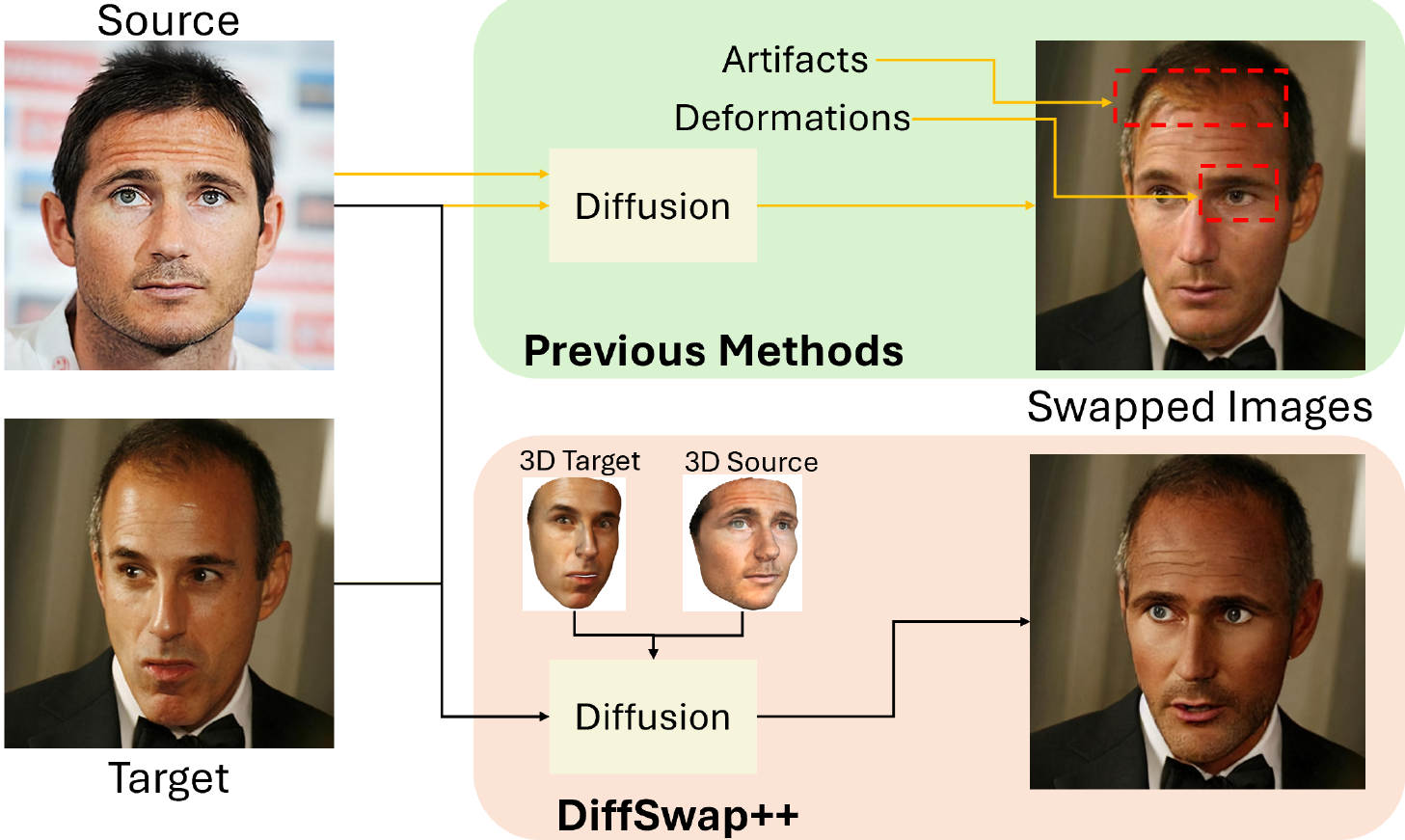}}
    \caption{Comparison between \textbf{Diffswap++} and previous diffusion methods showcasing the lack of artifacts and deformations in our swapped outputs.}
    \label{teaser}
    \vspace{-0.25in}
\end{figure}

While diffusion-based methods generate high-fidelity face swaps through multi-step processes using either DDPM~\cite{ddpm} or DDIM~\cite{ddim}, they remain prone to fine-grained artifacts. This limitation arises from their strong dependence on training data that sufficiently captures such challenging scenarios.
Unlike unconditional models, which generate data without explicit guidance, conditional diffusion models enable control over the generation process by incorporating auxiliary inputs such as text descriptions, class labels, or images~\cite{bao2022conditionalgenerativemodelsbetter}. The aforementioned face-swapping approaches such as DiffFace~\cite{diffface}, DiffSwap~\cite{diffswap}, and REFace~\cite{reface} leverage conditional diffusion frameworks using different identity conditioning strategies including identity embeddings, landmark features, or pretrained CLIP features~\cite{clip}. However, these methods do not explicitly incorporate the underlying 3D facial structure, which can be critical for achieving robust, controllable, and photorealistic face-swapping results.
Although DiffSwap utilizes 3D facial features, their use is limited to incorporating 3D face landmarks at inference and not directly employed to guide the training process.

To address the lack of 3D awareness in diffusion-based face-swapping approaches, we propose \textbf{DiffSwap++}, a novel pipeline that incorporates 3D facial information during training, yet operates without any 3D features at inference time. 
%CHANGES BELOW
By integrating 3D latent features during training, our method achieves enhanced visual realism and improved levels of identity preservation. As shown in figure \ref{teaser} This 3D supervision enables the model to more effectively disentangle facial identity from pose and expression.
Specifically, we leverage a 3D face reconstruction model~\cite{eg3d} to project 2D face images into 3D latent embeddings, which are then rendered back into 2D views. These 3D-aware latent features, capturing the structural layout of the face, are used to guide the diffusion model during training. Our proposed diffusion pipeline formulates face swapping as a target inpainting task, conditioned on both facial landmarks and identity embeddings. 

Through extensive experiments on the CelebA~\cite{celebA}, FFHQ~\cite{ffhq}, and CelebV-Text~\cite{celebvtext} datasets, we find that our model consistently outperforms existing methods in preserving the identity of the source face during swapping, while also maintaining comparable performance in retaining the structural attributes of the target face. 
Additionally, inspired by evaluation practices in biometric recognition research~\cite{biometrics1, biometrics2}, we assess the identity preservation capabilities of our model and several baselines by analyzing the trends between Impostor Attack Presentation Acceptance Rate (IAPAR) and the False Reject Rate (FRR) across varying acceptance thresholds. This biometric-style evaluation provides a more rigorous and quantitative understanding of how effectively each method retains the source identity during face swapping.

Our main contributions are summarized as follows:
\begin{itemize}
    \item We propose \textbf{DiffSwap++}, the first face-swapping framework that leverages 3D facial latent features to guide a diffusion-based generation pipeline. Our method incorporates 3D geometric cues during training while remaining fully 2D at inference.
    
    \item We conduct extensive experiments on three publicly available datasets that demonstrate the superior identity preservation and structural consistency of our approach. Additionally, a user study confirms that DiffSwap++ produces more realistic and visually convincing face swaps compared to prior methods.
    
    \item We introduce a novel evaluation perspective for face-swapping models by analyzing trends between IAPAR and False Reject Rate FRR, offering a rigorous biometric analysis of identity preservation performance.
\end{itemize}

%------------------------------------------------------------------------
\section{Related Work}
\label{sec:related}

\subsection{GAN Face-Swapping Approaches}
Generative Adversarial Networks, GANs are a class of deep learning models that can generate realistic images~\cite{gans}. Earlier approaches to the face-swapping task utilized GANs, training them to swap the attributes of a source face image onto a target face image while preserving the pose and expression of the face in the target image~\cite{simswap, hififace, faceshifter, hireslatent}. Despite promising results from many of these GAN based approaches to face-swapping, these models continued to experience issues inherent to GAN models such as a lack training stability and requiring careful hyperparameter tuning. These approaches also largely failed to produce high fidelity results that are often desired when performing face-swapping. More recently diffusion based approaches to face swapping have been introduced to remedy these issues.

\subsection{Diffusion Face-Swapping Approaches}
Diffusion models offered the training stability and high fidelity that GANs lacked, showing promising improvements on image synthesis~\cite{diffusionovergans}. The first face-swapping framework utilizing a diffusion approach came in DiffFace~\cite{diffface}. The DiffFace framework proposes the training of an ID Conditional DDPM~\cite{ddpm}, the output of which requires multiple stages of guidance at inference and needs to be progressively blended onto the original target face. DiffSwap~\cite{diffswap} posed face-swapping as an inpainting task where only the central portion of the face is considered as a candidate for swapping, eliminating the need for any blending at inference. REFace~\cite{reface} leverages features from an image foundation model (CLIP~\cite{clip}) in the face-swapping pipeline to provide the model with a semantic understanding of both the source and target face images. These methods lead to face-swaps which either contain artifacts or make only subtle modifications to the target faces and fail to transfer the identity of the source face effectively. These issues can be minimized by utilizing 3D features to disentangle face identity from pose and expression.

\subsection{3D Awareness in Diffusion Models and Face-Swapping Approaches}
Previous works have integrated 3D awareness in 2D diffusion processes~\cite{mvdiffusion, 3dfuse, genvs}. MVDiffusion~\cite{mvdiffusion} proposes a method for multi-view image generation by integrating camera pose conditioning and enforcing geometry constraints across latent representations. The diffusion process is further guided by cross-view attention, ensuring that image outputs remain consistent with underlying 3D scene structure. 3DFuse~\cite{3dfuse} conditions a pretrained 2D diffusion model using rendered depth maps from a coarse point cloud generated with a text prompt. This 3D structure is injected during sampling, allowing the 2D images to better reflect realistic shapes and poses. GeNVS~\cite{genvs} utilizes a learned latent 3D feature volume derived from a single source image. This latent space can then be rendered into novel views and the result is used to condition a 2D diffusion model, allowing for 3D-aware novel view synthesis with 2D outputs. 

Face-swapping methods have also attempted to implement 3D features. Some GAN based models utilize 3D features in their face-swapping pipelines. HiFiFace~\cite{hififace} generates 3D reconstructions of both the source and target faces, utilizing the shape information provided by these reconstructions for guidance during training. 3dSwap~\cite{3dswap} projects 2D faces into the 3D latent space where face-swapping is performed before the 3D swapped face is then rendered back to 2D. However, this method requires camera position information at inference to perform face swapping so the results of this model are notably absent from our comparison. The only diffusion-based approach to the face-swapping task that is 3D aware is DiffSwap~\cite{diffswap}. However, in DiffSwap the use of 3D features is rather limited as 3D face reconstruction is only used to generate landmarks that are provided as input to the UNet at inference. Our proposed method brings 3D awareness to the task of diffusion-based face-swapping in a much more substantial way as we utilize informative 3D latent features that are incorporated as conditioning features during the training stage. \modelname~does not require 3D features at inference.

%------------------------------------------------------------------------
\section{Proposed Method: DiffSwap++}
\label{sec:method}

\subsection{Preliminaries on Diffusion Models}
\label{subsec:diffusion_prelims}

To mitigate the computational expense of pixel-space generation, \modelname~operates entirely within the semantic latent space of a pretrained Latent Diffusion Model (LDM) \cite{latent}. Furthermore, to improve training and sampling efficiency over standard Denoising Diffusion Probabilistic Models (DDPMs) \cite{ddpm}, we build upon Denoising Diffusion Implicit Models (DDIMs) \cite{ddim}, which abandon the Markov assumption in favor of a deterministic update to predict the fully denoised representation $\hat{z}_0$ at each step.

\subsection{Face-Swapping Task}
To formulate the face-swapping task, given a source image $x^s$ and a target image $x^t$, the goal is to generate a new image in which the facial identity matches that of the source image $x^s$, while preserving the pose, expression, and background of the target image $x^t$.
In our proposed \modelname, we train a diffusion model for conditional inpainting, where the region consisting of the face of the target image is masked, and the face is reconstructed using conditioning features in the diffusion process. These conditioning features include the identity features of $x^s$ and characteristic features of $x^t$. Along parallel lines of research, ReFace~\cite{reface} introduces CLIP~\cite{clip} features of both $x^s$ and $x^t$ as conditioning. However, these features fail to capture the fine-grained expressions of the source face $x^s$ and the general pose of the target face $x^t$ and hence result in swaps which lack photorealism. To mitigate this, we introduce 3D latents into the diffusion model to guide the generative process in maintaining the pose and expression of the target face $x^t$ while preserving the identity of the source face $x^s$.

\iffalse
\subsection{Diffusion Input and Augmentation}
Following from past works in diffusion-based face-swapping we train our diffusion model to fill only a masked inner portion of the face, maintaining the rest of the features of the target image outside of this masked portion \cite{diffswap, reface}. As input to our DDIM UNet during training we provide our noised target face image at timestep \textit{$t$}, \textit{$X'_{\substack{tar \\ t}}$}, our masked target face \textit{$X_{inp}$}, and the mask used for our target face \textit{$M_{tar}$}. Alongside \textit{$X'_{\substack{tar \\ t}}$}, \textit{$X_{inp}$}, and \textit{$M_{tar}$} we provide as input a conditioning feature \textit{$f$}, which guides our diffusion model to generate high quality face swaps. To generate the conditioning feature we provide the target face image \textit{$X_{tar}$} alongside an augmented face image \textit{$X_{aug}$}. This \textit{$X_{aug}$} is derived from a face image \textit{$X$} where only the masked facial region remains and one or more of the following augmentations may have been randomly applied: horizontally flipping the face, rotating the face up to 20 degrees, slightly blurring the face, or elastically distorting the face. Performing this random augmentation of the source face \textit{$X_{src}$} adds diversity to the training data, leading to a more robust final model.
\fi

\begin{figure}
    \centering
    \includegraphics[width=1\linewidth]{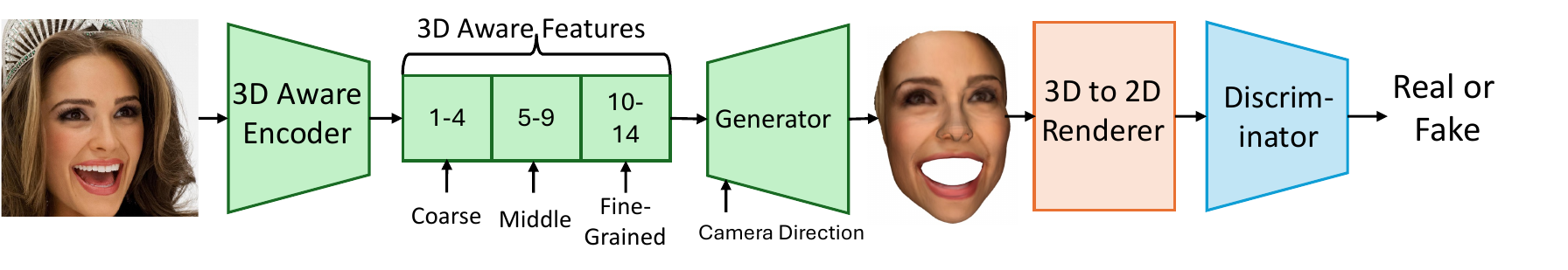}
    \caption{The 3D features are extracted from an encoder trained for 3D face reconstruction and are projector into a compatible feature dimension to be used as a conditioning feature in \modelname.}\vspace{-0.2in}
    \label{3D_features}
\end{figure}

\begin{figure*}
    \centering
    \scalebox{0.7}{
    \includegraphics[width=1.25\linewidth]{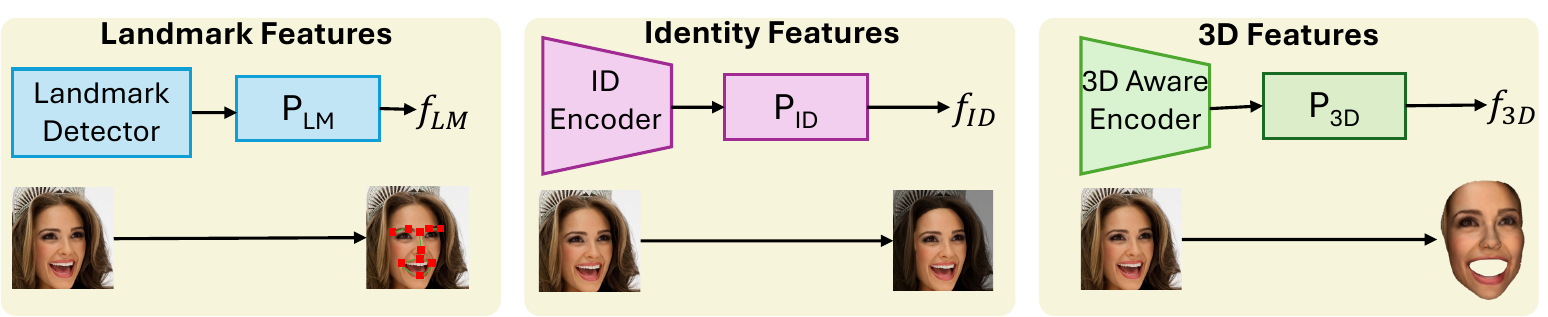}
    }
    \caption{Overview of the condition generation pipeline for \textbf{DiffSwap++}. We utilize the standard landmark and identity features alongside our 3D features, novel to diffusion face-swapping.} \vspace{-0.2in}
    \label{fig:condition_gen}
\end{figure*}

\subsection{Conditioning for Diffusion Guidance}
% As mentioned previously we utilize a conditioning feature \textit{$f$} to guide our diffusion process. This conditioning feature includes identity embeddings generated using ArcFace \cite{arcface} and key facial landmarks, both of which are standard in diffusion face-swapping approaches \cite{diffface, diffswap, reface}, giving us \textit{$f_{\text{ID}}$} and \textit{$f_{LM}$} respectively. 
For effective face-swapping, it is essential for the generated image to preserve (i) the identity of the source image $x^s$ and (ii) pose and expression of the target image $x^t$.
To obtain the identity features $f_{\text{ID}}$ of $x^s$, latent representations of a pretrained Face Recognition model $F_{\text{ID}}$ (e.g. ArcFace~\cite{arcface}, CosFace~\cite{cosface}) are projected using an MLP $P_{\text{ID}}$:

\begin{equation}
  f_{\text{ID}} = P_{\text{ID}}(F_{\text{ID}}(x^s)).
\end{equation}

To extract pose and expression features, the facial landmarks obtained from a landmark extractor $F_{\text{LM}}$ are projected using another MLP $P_{\text{LM}}$ to obtain the landmark feature representation of the target face $x^t$:

\begin{equation}
  f_{LM} = P_{LM}(F_{LM}(x^t)).
\end{equation}
While these landmark features capture 2D information about the pose and expression of the target face, they are often insufficient in challenging scenarios where the pose or expression deviates significantly from that of the source image. In such cases, diffusion models struggle to generate realistic and coherent face swaps. To address these challenges, \modelname~incorporates 3D latent representations of both the source and target images to more effectively guide the diffusion process, with the extraction of these 3D features depicted in Fig.~\ref{3D_features}). Our overall conditioning process can be viewed in Fig.~\ref{fig:condition_gen}.

\textbf{3D Conditioning Latent.} In \modelname, we incorporate 3D latent features as conditioning vectors for cross-attention with the Diffusion UNet. These 3D latents are generated by leveraging the encoder, $\mathcal{E}$, from a pretrained GAN model. During training the encoder is used to project an input image $x$ into latent space $\mathcal{W}$ to get a 3D aware high-dimension intermediate latent vector ${w_x}$. Then the pretrained EG3D\cite{eg3d} generator $\mathcal{G}$ is used to generate a triplane 3D representation of $x$ before rendering this representation back into a final 2D reconstruction $x' = \mathcal{G}(w_x, d)$ where $d$ is the input camera direction of $x$ estimated by Deep3d Face Reconstruction\cite{deepface3drecon}. Additionally, the EG3D generator is used to generate another view of the input face $\hat{x} = \mathcal{G}(w_x, \hat{d})$ where $\hat{d}$ is a random camera direction. This new view of the input face is then fed into the encoder to get $w_{\hat{x}}$ and the reconstruction process is repeated once again to produce $\hat{x}' = \mathcal{G}(w_{\hat{x}}, d)$. The overall optimization of this model can be written as: $\min_{\theta}\{\mathcal{L}(x, x') + \eta\mathcal{L}(x, \hat{x}') + \mathcal{L}(w_x, w_{\hat{x}})\}$. The latent representations generated by this encoder are of a standard dimension for 3D face representations consistent with that of EG3D\cite{eg3d}.
% The layers within these latent representations represent different granularities of attributes. The latent space is comprised of 14 layers, the layers 1-4 denote the ``\textit{coarse}" features such as pose, general hair style, face shape, and eyeglasses; layers 10-14 representing the ``\textit{fine-grained}" features such as the general color scheme and background; and layers 5-9 representing the ``\textit{middle}" features such as smaller scale facial features, hair style, and if the eyes are open or closed. The 3D latent $f_{\text{3D}}$ captures this complementary information by aggregating the information of all the feature layers. Thus, 
\begin{figure}
    \centering
    \scalebox{0.6}{
    \includegraphics[width=1.25\linewidth]{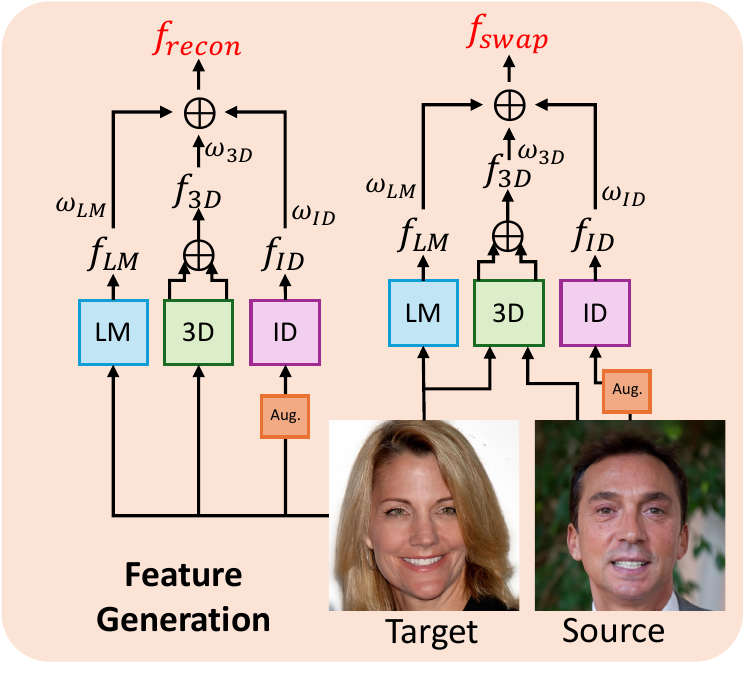}
    }
    \caption{Our proposed \textbf{feature generation pipeline}. The obtained \textcolor{red}{features} are fed as conditional input to our Diffswap++ diffusion model.}
    \label{fig:main_arch2}
\end{figure}

The latent space is hierarchically organized by feature frequency, inherited from the underlying 3D generator structure. We categorize these into three levels of granularity to analyze their specific impacts:
\begin{itemize}
    \item \textbf{Coarse-grained (Layers 1-4):} These layers control low-frequency structural geometry, including head shape, volumetric features, and pose orientation. They provide the fundamental geometric ``canvas'' of the source identity.
    \item \textbf{Middle-grained (Layers 5-9):} These layers govern localized facial features and semantic attributes, such as eye shape, mouth structure, and specific expression nuances.
    \item \textbf{Fine-grained (Layers 10-14):} These layers represent high-frequency details, including skin texture, lighting conditions, and color schemes.
\end{itemize}
The final 3D latent $f_{3D}$ captures the holistic structural identity by aggregating these complementary frequencies. Thus,
\vspace{-0.95em}
\begin{align}
  f_{\text{3D}}^t &= P_{\text{3D}}^t(\frac{1}{\text{L}}\sum_{i=1}^{\text{L}}F_{\text{3D}}(x^t)), \\
  f_{\text{3D}}^s &= P_{\text{3D}}^s(\frac{1}{\text{L}}\sum_{i=1}^{\text{L}}F_{\text{3D}}(x^s)), \\
  f_{\text{3D}} &= f_{\text{3D}}^t + f_{\text{3D}}^s 
\end{align} 
Here, $P_{\text{3D}}^t$ and $P_{\text{3D}}^s$ are MLPs to project the 3D latents of the source and target images and L is the number of layers of the latent features. 
Finally, all latent features are then combined to create the final conditioning feature \textit{$f$} as follows
\vspace{-0.3em}
\begin{equation}
f = w_{\text{3D}}f_{\text{3D}} + w_{\text{ID}}f_{\text{ID}} + w_{LM}f_{LM}\label{eq:9} \vspace{-0.5em}
\end{equation}
where $w_{\text{3D}}$, $w_{\text{ID}}$ and $w_{LM}$ are weights for aggregation as illustrated in Figure~\ref{fig:main_arch2}.

\begin{figure}
    \centering
    \scalebox{0.7}{
    \includegraphics[width=1.35\linewidth]{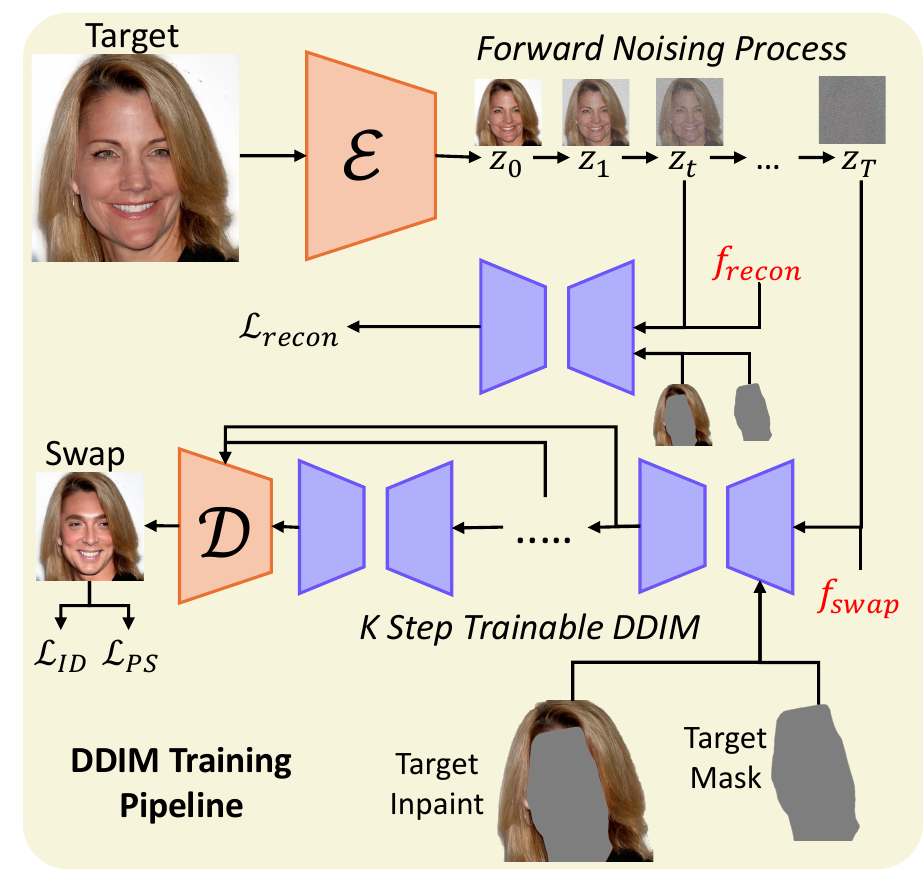}
    }
    \caption{Training pipeline for \textbf{DiffSwap++}. This figure highlights our primary DDIM training pipeline where we perform both reconstruction and face-swapping.}\vspace{-0.1in}
    \label{fig:main_arch1}
\end{figure}

\subsection{Training Objective of \modelname} %TODO
Our training objective fuses three complementary losses as illustrated in Fig~\ref{fig:main_arch1}, viz., denoising diffusion, identity preservation, and perceptual similarity, optimized in two sequential steps.  In the \textit{face-reconstruction step}, a U-Net denoiser $\epsilon_{\theta}$ learns to infer the Gaussian noise $\epsilon$ injected at a random diffusion timestep $t$.  Given the noisy sample $z_{t}$, its timestep embedding $x^{t}$, a spatial mask $M^{tar}$, and our fully conditioned 3D aware feature token $f_{recon}$ distilled from the target image, the network is supervised with the mean–squared error: 
\begin{equation}
\mathcal{L}_{\text{Recon}} = E_{t,x_0,\epsilon} \left\| \epsilon_{\theta} (\textit{$z_t$},M^{tar}, f_{recon}, t) - \epsilon \right\|_2^2
\end{equation}
The \textit{face–swapping step} fine-tunes a DDIM sampler, so that its intermediate latents $\hat{z}_{i} (i\!=\!1 \dots N)$ transmit the source identity while blending seamlessly with the target context. To calculate our losses the intermediate latents are processed through a decoder $\mathcal{D}$ and, for $\mathcal{L}_{\text{ID}}$, an ArcFace~\cite{arcface} identity encoder $\mathcal{E}$:
\begin{equation}
\mathcal{L}_{\text{ID}} = \sum_{i=1}^{\text{N}} \left(1 - \left< \mathcal{E}(\mathcal{D}(\hat{z}_{0(z_i, t_i)}) \otimes M^t), \mathcal{E}(x^s \otimes M^s) \right> \right)
\end{equation}
where $M^s$ and $M^t$ are source and target face masks respectively and $\langle\!\cdot,\!\cdot\rangle$ denotes cosine similarity.  Visual coherence with the target background is promoted via a learned perceptual image patch similarity (LPIPS) metric\cite{LPIPS},

\begin{equation}
\mathcal{L}_{\text{PS}} = \sum_{i=1}^{\text{N}} L_{\text{LPIPS}} (\mathcal{D}(\hat{z}_{0(z_i, t_i)}), x^t)
\end{equation}
The total loss \textit{$\mathcal{L}_{\text{Total}}$} is thus a` combination of reconstruction, identity and perceptual similarity losses denoted as \begin{equation}
\mathcal{L}_{\text{Total}} = \mathcal{L}_{\text{Recon}} + w_{\text{LID}}\mathcal{L}_{\text{ID}} + w_{\text{PS}}\mathcal{L}_{\text{PS}}\label{eq:13}
\end{equation}
where $w_{\text{ID}}$ and $w_{\text{PS}}$ trade off identity preservation against perceptual realism.  This two-step optimization first instills a strong generative prior through accurate face reconstruction, then aligns the model with the face-swapping objectives without sacrificing realism or identity fidelity.

\section{Experiments}
\label{sec:exp}
\noindent
\textbf{Datasets.} \modelname~is trained on the CelebA-HQ~\cite{celebahq} dataset where the images are of high quality version of the CelebA~\cite{celebA} dataset. The dataset comprises of 30k images, of resolution $1024 \times 1024$, of which 28k are used for training and the remaining for evaluation. In addition, for out-of-domain (OOD) data, we use the FFHQ dataset \cite{ffhq} for evaluation. FFHQ contains 70k high quality face images and we utilize the last 2k images for our evaluation. 
For biometric evaluations for source identity consistency in swapped faces, we use the CelebV-Text~\cite{celebvtext} dataset. We sample 50 frames from each of 20 videos from the dataset to perform face-swapping.

\noindent
\textbf{Implementation Details.} 
In all our experiments, the images are resized to $512 \times 512$. The denoising network ($\epsilon_{\theta}$) is a U-Net architecture~\cite{unet}. For the 2D conditioning features, ID features are extracted using ArcFace~\cite{arcface}. We obtain 68 face landmark positions from DLib~\cite{dlib}. The 3D conditioning feature latents are generated by leveraging an encoder with a feature pyramid architecture from pSp~\cite{psp} that is pre-trained to generate 3D latent representations of 2D faces in 3DSwap~\cite{3dswap}. These conditioning features are integrated into the diffusion process using cross attention~\cite{cross} with each U-Net layer.

Our model is trained on 4 NVIDIA A6000 GPUs (48GB). We train our model for 20 epochs with a global batch size of 4 and initialize our training for stable diffusion \cite{latent}. During training, we perform $T=1000$ noising time-steps and $N=4$ DDIM~\cite{ddim} steps. During inference, $50$ DDIM~\cite{ddim} steps are used. For the weighted sum in equation \ref{eq:9} to calculate $f$, we utilize the hyperparameters $w_{\text{3D}}=1.0$, $w_{\text{ID}}=10.0$, and $w_{\text{LM}}=0.05$. For the weighted sum in equation \ref{eq:13} to calculate \textit{$\mathcal{L}_{\text{Total}}$}, we utilize the hyperparameters $w_{\text{ID}}=0.3$ and $w_{\text{PS}}=0.1$.

\noindent
\textbf{Evaluation Protocol.} 
We benchmark the face-swapping performance of \modelname~on the CelebA-HQ~\cite{celebahq} and FFHQ~\cite{ffhq} test sets. Specifically, each method is applied to an identically constructed set of 1000 source faces and 1000 target faces, yielding 1000 source–target pairs and ensuring strictly comparable conditions across all baselines. To assess the potential security risks, we report Fr\'echet Information Distance (FID) \cite{fid} to indicate the difficulty of detecting generated samples compared to real data. We measure impersonation potential using top-1 and top-5 retrieval accuracy, which proxies the success rate of bypassing identity verification systems. For the identity metric, ArcFace \cite{arcface} is used to embed every source and swapped face, after which cosine similarity determines the nearest neighbors among the source set. Pose and expression measured as the $\ell_{2}$ distance between head-pose estimates from HopeNet \cite{hopenet} and expression coefficients from DeepFace3DRecon \cite{deepface3drecon} are evaluated to ensure the contextual consistency necessary to deceive human observers in dynamic environments. All baseline methods are evaluated using their officially released code and publicly available pretrained checkpoints.
Note that the results for ReFace~\cite{reface} are obtained from a model we trained for 20 epochs using the exact hyperparameter settings reported in the original paper, to ensure a fair comparison.
\vspace{-0.15in}

\begin{table}[h]
\centering
\caption{Inference time and memory usage per image.}
\label{efficiency}
\small
\begin{tabular}{llcc}
\toprule
\textbf{Method} & \textbf{Type} & \textbf{Time (s)} & \textbf{VRAM (GB)} \\
\midrule
DiffFace & Diffusion & 37.77 & 8.15 \\
DiffSwap & Diffusion & 5.20  & 8.07 \\
FaceDancer & GAN & 0.80  & 1.65 \\
HiFiFace & GAN & 0.04  & 2.80 \\
REFace & Diffusion & 3.44  & 18.68 \\
\rowcolor{lightblue} \textbf{\modelname} & \textbf{Diffusion} & \textbf{3.35} & \textbf{18.69} \\
\bottomrule
\end{tabular}
\vspace{-0.15in}
\end{table}

\textbf{Inference Efficiency.} Table \ref{efficiency} compares the computational efficiency of our method against state-of-the-art baselines. In terms of inference speed, DiffSwap++ requires $3.35$s per image at inference, substantially faster than diffusion-based baselines: $\sim11\times$ faster than DiffFace and $\sim1.5\times$ faster than DiffSwap, while remaining on par with REFace. In terms of memory, DiffSwap++ uses $18.69$\,GB VRAM. Although higher than lightweight methods such as FaceDancer~\cite{facedancer}, it is comparable to high-fidelity frameworks and fits within the capacity of modern GPUs.
 
\begin{table}[t]
\centering
\caption{Comparison on CelebA~\cite{celebA} dataset.}
\label{CelebA_Quant}
\setlength{\tabcolsep}{3pt}
\resizebox{\columnwidth}{!}{%
\begin{tabular}{lccccc}
\hline
\multirow{2}{*}{\textbf{Method}} & \multirow{2}{*}{\textbf{FID}$\downarrow$} & \multicolumn{2}{c}{\textbf{ID retrieval} $\uparrow$} & \multirow{2}{*}{\textbf{Pose}$\downarrow$} & \multirow{2}{*}{\textbf{Expr.}$\downarrow$} \\
& & \textbf{Top-1} & \textbf{Top-5} & & \\
\hline
HiFiFace~\cite{hififace} & 13.97 & 86.4\% & 93.9\% & 3.01 & 1.19 \\
FaceDancer~\cite{facedancer} & 12.25 & 79.0\% & 92.9\% & 2.32 & \textbf{0.65} \\
DiffFace~\cite{diffface} & 9.49 & 93.0\% & 97.2\% & 3.19 & 1.01 \\
DiffSwap~\cite{diffswap} & 8.29 & 16.9\% & 34.9\% & \textbf{2.13} & 0.76 \\
SimSwap~\cite{simswap} & 10.61 & 90.3\% & 93.3\% & 2.21 & 0.78 \\
REFace~\cite{reface} & 7.01 & 92.7\% & 96.9\% & 3.30 & 1.01 \\ \hline
DiffSwap++ (w/o 3D) & 8.80 & 86.0\% & 94.1\% & 3.48 & 1.06 \\
DiffSwap++ (NeRF Feat.) & 9.64 & 92.0\% & 96.9\% & 4.16 & 1.12 \\
DiffSwap++ (e4e Enc.) & 7.24 & 91.4\% & 97.0\% & 3.94 & 1.05 \\
\rowcolor{lightblue} \textbf{\modelname} & \textbf{6.42} & \textbf{94.7}\% & \textbf{98.2}\% & 3.29 & 1.00 \\
\hline
\end{tabular}%
} \vspace{-0.2in}
\end{table}

\subsection{Comparison with the State-of-the-Art}
\noindent
\textbf{Face-Swapping on CelebA-HQ and FFHQ.} 
%MODIFIED TO INCLUDE REFERENCE TO BASELINE
In Table~\ref{CelebA_Quant}, we compare \modelname~against state-of-the-art face-swapping methods, as well as a variant of \modelname~trained without 3D features, using 1{,}000 source--target pairs from CelebA-HQ~\cite{celebahq}. In Table~\ref{FFHQ_Quant}, we further evaluate \modelname~under out-of-domain (OOD) face swapping on the challenging FFHQ~\cite{ffhq} dataset. Across both datasets, \modelname~achieves the highest top-1/top-5 identity retrieval accuracy and the lowest Fr\'echet Information Distance (FID), demonstrating that it produces swaps that are both photorealistic and highly effective at inducing false acceptances in recognition systems. Importantly, pose and expression similarity remain competitive with the strongest baselines, indicating that improved identity fidelity is not obtained at the expense of geometric or affective consistency.

We note that identity retrieval is inherently at odds with pose and expression preservation: methods that minimally modify the target face typically achieve strong pose/expression scores but fail to transfer the source identity. Consequently, approaches reporting low identity retrieval often exhibit favorable pose and expression metrics, as they avoid the structural changes required to preserve the source identity in the swapped result.
This trade-off is reflected in our results: DiffSwap~\cite{diffswap}, SimSwap~\cite{simswap}, and FaceDancer~\cite{facedancer} largely preserve the target pose and expression, yielding strong scores on those metrics, but their limited identity transfer leads to higher FID and substantially lower identity-matching performance. In contrast, \modelname~strikes a more balanced compromise, achieving state-of-the-art identity preservation while maintaining the target's pose and expression with minimal degradation.

In Table~\ref{CelebA_Quant} and Table~\ref{FFHQ_Quant}, we further observe that our 3D-aware latent features are more effective for training than alternative 3D feature extraction strategies. Specifically, we evaluate (i) replacing our 3D features with NeRF features extracted using a pi-GAN~\cite{pigan} model pretrained on CelebA~\cite{celebA}, and (ii) substituting the pSp~\cite{psp} encoder in our 3D pipeline with an e4e~\cite{e4e} encoder. Although both variants converge faster (8 epochs for e4e and 10 epochs for NeRF) and improve ID retrieval, further supporting the robustness of our 3D guidance, their overall performance remains inferior to DiffSwap++.

\noindent
\textbf{Biometric Evaluation for Deepfake Detection.} To gauge how faithfully a face swapping method propagates source identity under a biometric lens, it is natural to analyze the Impostor Attack Presentation Acceptance Rate (IAPAR) metric across different acceptance thresholds \cite{biometrics1, biometrics2}. Biometric security systems use IAPAR to estimate their effectiveness in identifying fraudulent faces instead of incorrectly accepting them as genuine users. We utilize IAPAR to strictly evaluate the impersonation success of the swap acting as a presentation attack, distinguishing this metric from standard detection metrics like AUC.

\begin{figure}
    \centering
    \scalebox{0.8}{
    \includegraphics[width=1\linewidth]{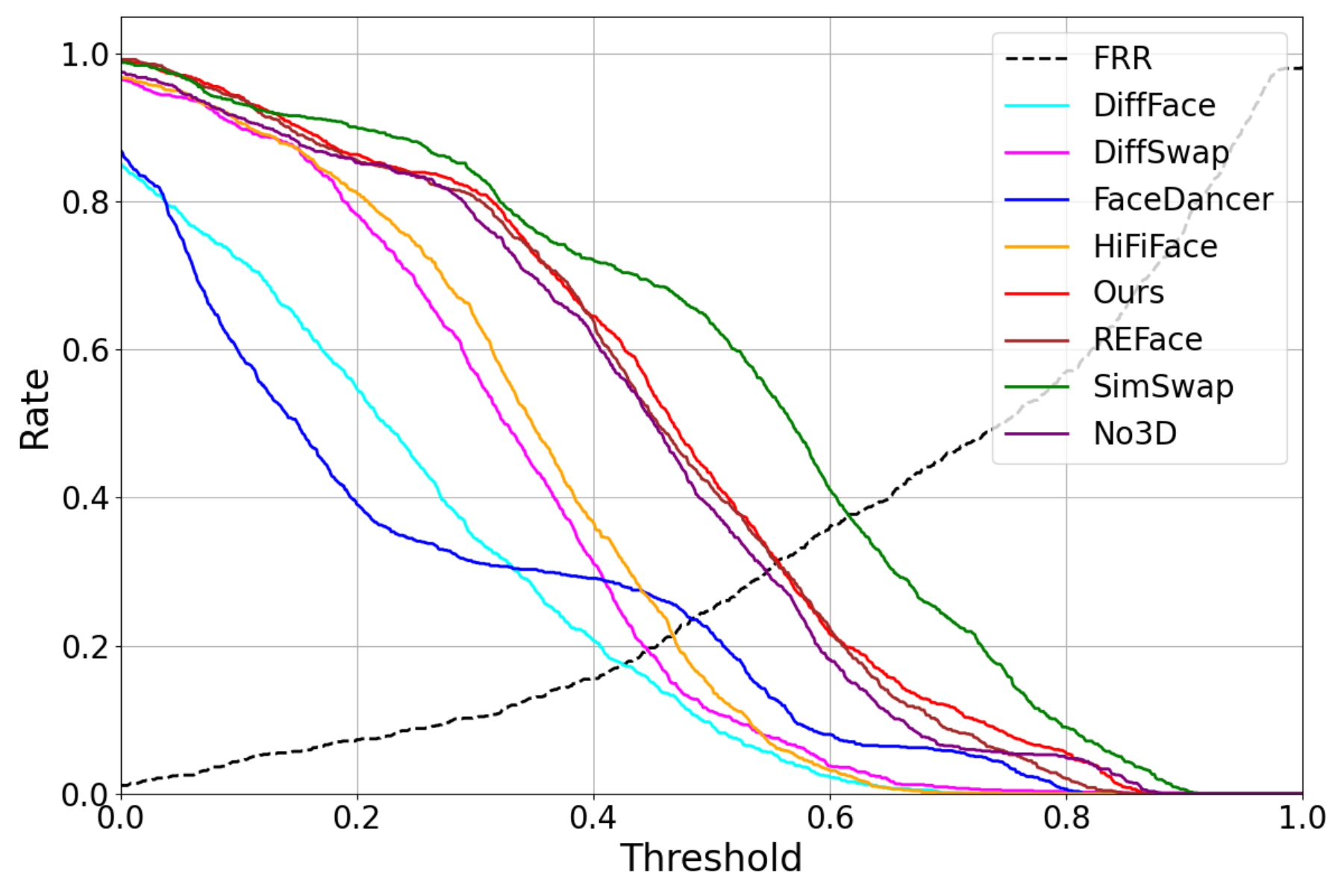}}
    \caption{A comparison of IAPAR for all models at different thresholds against FRR.} \vspace{-0.1in}
    \label{IAPAR}
\end{figure}

\begin{figure}
    \centering
    \scalebox{0.8}{
    \includegraphics[trim=0cm 3.7cm 0cm 0cm, clip, width=1\linewidth]{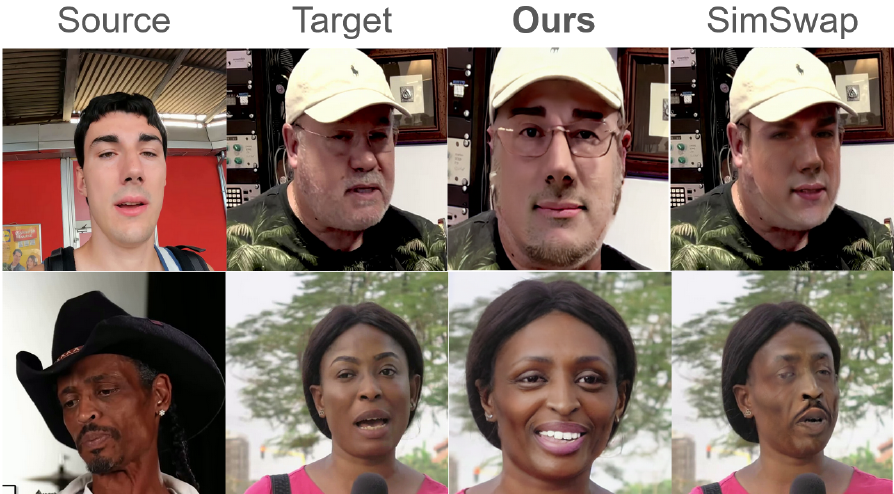}}
    \caption{Comparison between \textbf{DiffSwap++} and SimSwap on CelebV-Text video frames.}  
    \label{vtext_qual}
    \vspace{-0.15in}
\end{figure}

For each model, we construct a face-swap corpus on CelebV-Text \cite{celebvtext} by randomly selecting 20 videos, sampling 50 frames per video as targets. We replace each target face with a single source frame drawn from a different video, yielding 1000 swapped frames per method. FaceNet-512~\cite{facenet} embeddings are extracted from every swapped frame and its corresponding source image, and cosine similarity scores are thresholded over the full $[0, 1]$ range; IAPAR is then the fraction of swapped frames whose similarity exceeds the threshold, i.e., the rate at which an imposter presentation (the swap) is incorrectly accepted as the genuine source. We compute the False Reject Rate (FRR) in an analogous way, but using 50 original frames from the same video as the source image. In Fig.~\ref{IAPAR}, we plot the percentage of ground truth frames that fail to match the identity of the corresponding source image at thresholds ranging from $0-1$. \modelname~maintains a high threshold demonstrating high-fidelity face-swaps that preserve the source face identity. SimSwap~\cite{simswap} demonstrates a higher threshold than \modelname. However, qualitative inspection in Fig.~\ref{vtext_qual} shows that SimSwap generates visually implausible swaps failing to properly integrate the source identity onto the target features, leading to swaps which are similar in identity to the source. On the other hand, \modelname~generates swaps that maintain realism.

\begin{table}[t]
\centering
\caption{Comparison on FFHQ~\cite{ffhq} dataset.}
\label{FFHQ_Quant}
\setlength{\tabcolsep}{3pt}
\resizebox{\columnwidth}{!}{%
\begin{tabular}{lccccc}
\hline
\multirow{2}{*}{\textbf{Method}} & \multirow{2}{*}{\textbf{FID}$\downarrow$} & \multicolumn{2}{c}{\textbf{ID retrieval} $\uparrow$} & \multirow{2}{*}{\textbf{Pose}$\downarrow$} & \multirow{2}{*}{\textbf{Expr.}$\downarrow$} \\
& & \textbf{Top-1} & \textbf{Top-5} & & \\
\hline
HiFiFace~\cite{hififace} & 11.85 & 74.0\% & 86.3\% & 3.20 & 1.39 \\
FaceDancer~\cite{facedancer} & 9.45 & 75.9\% & 87.6\% & 2.43 & \textbf{0.70} \\
DiffFace~\cite{diffface} & 8.72 & 86.9\% & 94.0\% & 3.67 & 1.30 \\
DiffSwap~\cite{diffswap} & \textbf{6.21} & 34.9\% & 55.7\% & 2.43 & 1.03 \\
SimSwap~\cite{simswap} & 9.48 & 77.8\% & 81.7\% & \textbf{2.39} & 0.92 \\
REFace~\cite{reface} & 6.45 & 91.0\% & 95.5\% & 3.69 & 1.07 \\ \hline
DiffSwap++ (w/o 3D) & 9.95 & 84.2\% & 92.4\% & 3.56 & 1.09 \\
DiffSwap++ (NeRF Feat.) & 10.75 & 93.4\% & 98.1\% & 5.32 & 1.16 \\
DiffSwap++ (e4e Enc.) & 5.74 & 94.1\% & 97.6\% & 4.07 & 1.09 \\
\rowcolor{lightblue} \textbf{\modelname}  & 6.57 & \textbf{95.1}\% & \textbf{98.4}\% & 3.70 & 1.07 \\
\hline
\end{tabular}%
}
\end{table}

\subsection{Ablation Analysis}
To understand how the 3D conditioning latent should be incorporated into the face-swap network, we experiment with five integration schemes for the $14\times512$ feature tensor produced by the 3D Feature Extractor~\cite{eg3d}. The first strategy performs a global average over the layer axis, yielding a single 512-D descriptor (`\textit{Avg. Pool}'). The next three strategies pool only over specific layer groups that roughly correspond to `\textit{Coarse}' (layers 1–4), `\textit{Middle}' (5–9), and `\textit{Fine-Grained}' (10–14) geometry and appearance, respectively. Finally, we replace pooling with a learnable linear projection that selects a weighted combination of all 14 layers. Each variant is trained for five epochs under identical hyper-parameters, and its quantitative performance on FFHQ\cite{ffhq} is reported in Table~\ref{tab:combined_ablation}.
This table reveals the specific influence mechanisms of different granularity levels. 

\textbf{Impact on Identity:} The ``Coarse'' variant yields the lowest identity retrieval (44.30\%), indicating that while head shape is necessary, it is insufficient for robust recognition without textural details. The ``Fine-Grained'' variant performs better (64.40\%) by capturing skin texture, yet fails to achieve state-of-the-art results without the structural support of the coarse layers. The superior performance of ``Avg. Pool'' (79.30\%) confirms that identity is a composite of both structural geometry and textural appearance.

\textbf{Impact on Pose Consistency:} Notably, the ``Coarse'' variant results in higher pose error ($l_2$ distance of 3.88) compared to ``Fine-Grained'' (3.56). We attribute this to the entanglement of source pose within the coarse layers; relying exclusively on coarse source latents creates a geometric conflict with the target's pose mask. By utilizing the full hierarchy, ``Avg. Pool'' balances structural identity transfer while mitigating these geometric conflicts, as evidenced by its robust performance across all metrics.

We additionally provide results for variants of our model, each trained for five epochs, in Table~\ref{tab:combined_ablation} where it is evident that 4 DDIM steps provide the best balance during training and that both identity and landmark features are required for training a face swapping model of this nature. 

\begin{table}[h!]
\centering
\caption{Ablation studies on 3D Condition Generation and DDIM Steps/Loss Components on FFHQ dataset.}
\label{tab:combined_ablation}
\setlength{\tabcolsep}{3pt}
\resizebox{\columnwidth}{!}{%
\begin{tabular}{lccccc}
\hline
\multirow{2}{*}{\textbf{Method}} & \multirow{2}{*}{\textbf{FID}$\downarrow$} & \multicolumn{2}{c}{\textbf{ID retrieval} $\uparrow$} & \multirow{2}{*}{\textbf{Pose}$\downarrow$} & \multirow{2}{*}{\textbf{Expr.}$\downarrow$} \\
& & \textbf{Top-1} & \textbf{Top-5} & & \\
\hline
\multicolumn{6}{l}{\textit{3D Condition Generation}} \\
\hline
\rowcolor{lightblue}\textbf{Avg. Pool} & \underline{5.72} & \textbf{78.80\%} & \textbf{90.20\%} & 4.40 & \textbf{1.04}  \\
Coarse & 6.59 & 46.90\% & 65.90\% & 4.11 & \textbf{1.04} \\
Middle & 7.21 & 54.20\% & 73.80\% & 4.12 & 1.09 \\
Fine-Grained & \textbf{5.68} & \underline{65.60\%} & \underline{79.80\%} & \textbf{3.85} & \textbf{1.04} \\
Linear Proj. & 6.27 & 50.90\% & 69.80\% & \underline{3.88} & \underline{1.05} \\
\hline
\multicolumn{6}{l}{\textit{DDIM Steps and Loss Components}} \\
\hline
\rowcolor{lightblue}\textbf{DiffSwap++} & \textbf{5.72} & \textbf{78.80\%} & \textbf{90.20\%} & \textbf{4.40} & \textbf{1.04}  \\
DiffSwap++ (w/o ID Feat.) & 15.84 & 0.10\% & 0.70\% & 5.88 & 1.15 \\
DiffSwap++ (w/o LM Feat.) & 42.19 & 0.30\% & 0.60\% & 20.84 & 2.54 \\
DiffSwap++ (2 DDIM Steps) & 9.28 & 32.30\% & 50.30\% & 5.58 & 1.27 \\
DiffSwap++ (6 DDIM Steps) & 27.7 & 11.40\% & 27.30\% & 4.91 & 1.40 \\
\hline
\end{tabular}%
}
\vspace{-0.2in}
\end{table}

\begin{figure*}[t]
    \centering
    \scalebox{0.9}{
    % Adjust the '2.5cm' value up or down until the bottom row is perfectly hidden
    \includegraphics[trim=0cm 7.5cm 0cm 0cm, clip, width=1\linewidth]{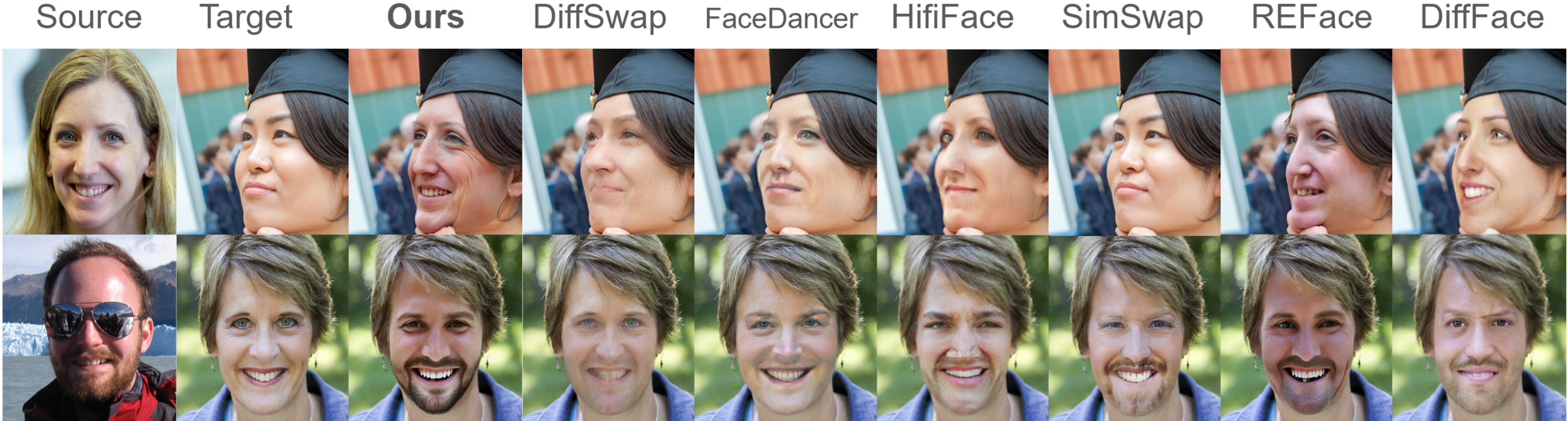}}
    \caption{Qualitative comparison of all models on FFHQ.} \vspace{-0.1in}
    \label{CelebA_Qual}
\end{figure*}

\begin{figure}[h!]
    \centering
    \scalebox{0.8}{
    \includegraphics[trim=0cm 3.15cm 0cm 0cm, clip, width=1\linewidth]{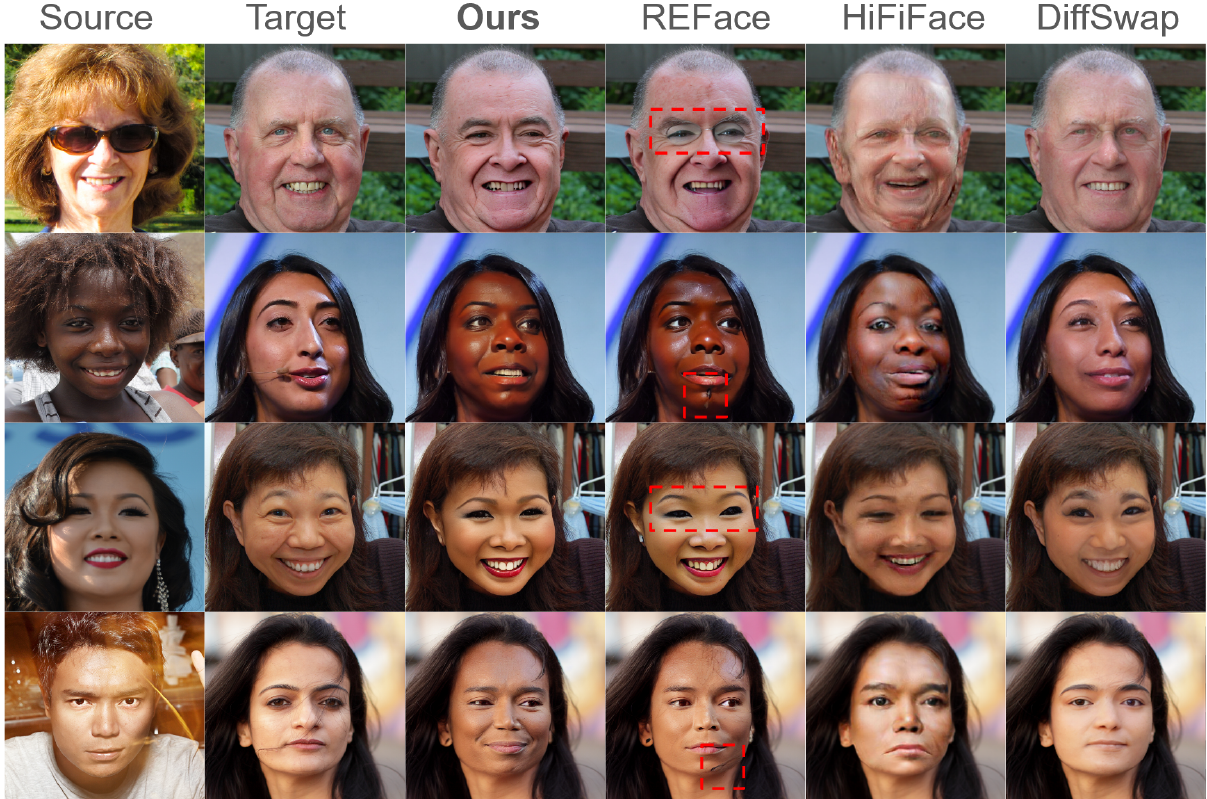}
    }
    \caption{Comparison highlighting issues with other models (artifacts highlighted in red).} 
    \label{select}
\end{figure}

\subsection{Qualitative Evaluation}
Qualitative comparisons between all models are provided in Fig.~\ref{CelebA_Qual}. Three recurrent failure modes observed in prior works are highlighted in Fig.~\ref{select} and compared to the face-swaps produced by \modelname. First, \textbf{visual artifacts} hallucinated eyeglass rims, hollow pupils, or irregular chin textures are prominent in the REFace~\cite{reface} outputs. These irregularities serve as clear indicators for forgery detection algorithms. In contrast, \modelname minimizes these cues, producing anatomically consistent geometry that significantly complicates the task of distinguishing synthetic content from authentic. Second, several methods synthesize \textbf{anatomically implausible faces} that break facial geometry; this is readily apparent in the face-swaps generated by HiFiFace~\cite{hififace}, but is absent in the ones by \modelname, whose structure adheres to plausible facial morphology. Third, approaches such as DiffSwap often \textbf{leave the target face intact}, effecting only superficial alterations; although this strategy yields high pose and expression scores, it fails to convey the source identity and therefore fares poorly on retrieval metrics. By contrast, \modelname~achieves a more balanced transformation, preserving identity while respecting target pose and expression, and doing so without introducing artifacts or geometric distortions.

\begin{table}[h!]  \vspace{-0.1in}
    \centering
    \caption{User Study Results} 
    \label{user}
    \resizebox{\linewidth}{!}{
        \begin{tabular}{l|l|c|c|c}
            \hline
            \textbf{Metric} & \textbf{Method} & \textbf{Mean} & \textbf{Var} & \textbf{95\% CI} \\
            \hline
            \multirow{4}{*}{\textbf{Identity}} 
            & \textbf{Ours} & \textbf{3.46} & 0.32 & [3.06, 3.86] \\
            & REFace & \underline{3.20} & 0.23 & [2.86, 3.54] \\
            & Hifi & 2.58 & 0.26 & [2.22, 2.94] \\
            & DS & 2.46 & 0.33 & [2.05, 2.87] \\
            \hline
            \multirow{4}{*}{\textbf{Expression}} 
            & Ours & 3.14 & 0.32 & [2.74, 3.54] \\
            & REFace & \underline{3.22} & \textbf{0.19} & [2.91, 3.53] \\
            & Hifi & 2.72 & 0.39 & [2.27, 3.17] \\
            & DS & \textbf{3.57} & 0.42 & [3.11, 4.03] \\
            \hline
            \multirow{4}{*}{\textbf{Pose}} 
            & Ours & \underline{3.65} & 0.29 & [3.26, 4.04] \\
            & REFace & 3.48 & \textbf{0.04} & [3.33, 3.63] \\
            & Hifi & 3.08 & 0.29 & [2.70, 3.46] \\
            & DS & \textbf{4.11} & 0.25 & [3.75, 4.47] \\
            \hline
        \end{tabular} 
    }
    \vspace{-0.1in}
\end{table}

\textbf{User Study for Face-Swap Evaluation.} To evaluate the risk of deceiving human moderators, we conducted a user study involving 10 volunteer participants. As human inspection remains a primary layer of defense in media verification, assessing the perceptual convincingness of the output is crucial. Each participant was presented with 10 sets of face-swapped images along with their corresponding source and target images. Each set included one face swap from \modelname, REFace~\cite{reface}, HifiFace~\cite{hififace}, and DiffSwap~\cite{diffswap}. 
Participants were asked to rate each face swap on three criteria: identity preservation, expression preservation, and pose preservation using a 5-point Likert scale, where 1 indicates the lowest quality and 5 indicates the highest. To eliminate bias, the order of face swaps within each set was randomized, and the names of the models were not disclosed. 
The quantitative results, summarized in Table~\ref{user}, demonstrate that \modelname~achieves the highest mean perceptual quality in identity preservation ($3.46$). While the performance gap with the strong baseline REFace ($3.20$) is not statistically significant ($p=0.26$), indicating comparable reliability between the two methods, the 95\% confidence interval for \modelname~($[3.06, 3.86]$) extends higher than that of REFace ($[2.86, 3.54]$), suggesting a higher upper bound for identity preservation. In contrast, our method significantly outperforms HifiFace ($p=0.006$) and DiffSwap ($p=0.014$), the latter of which sacrifices identity retention ($2.46$) for expression. Regarding stability, \modelname~maintains a balanced variance profile ($\sigma^2 \approx 0.32$) across all metrics. This stands in contrast to DiffSwap, which exhibits higher instability in expression ($\sigma^2=0.42$), and REFace, which, while highly stable ($\sigma^2=0.04$ in pose), fails to reach the high-fidelity peaks achieved by our method in both Identity and Pose. Thus, \modelname~offers the most effective balance of high mean fidelity and consistent generation quality.
\vspace{-0.1in}

\section{Conclusion}
\label{sec:conclusion}
In this work, we propose \modelname, a diffusion-based face-swapping pipeline that incorporates 3D facial features during training to effectively disentangle identity from pose and expression, enabling more accurate identity preservation in face-swapped images. Our approach is efficient at inference and achieves state-of-the-art performance in identity preservation, while maintaining comparable levels of pose and expression fidelity. We validate the effectiveness of our method through extensive experiments on three public datasets, including a biometric-style evaluation designed to rigorously assess identity preservation capabilities.

\section*{Acknowledgments}
This material is based upon work supported by the Center for Identification Technology Research and the National Science Foundation under Grant No. 1650503 and 2413228. We would also like to thank Kumara Kahatapitiya at Stony Brook University for our valuable discussions.

{\small
\bibliographystyle{ieee}
\bibliography{egbib}
}

% =================================================================
% SUPPLEMENTARY MATERIAL
% =================================================================
\clearpage
\twocolumn[
\begin{center}
    {\Large \bf DiffSwap++: 3D Latent-Controlled Diffusion for \\ Identity-Preserving Face Swapping \\ \vspace{0.5em} Supplementary Material \par}
    \vspace{1.5em}
    {\large Weston Bondurant, Arkaprava Sinha, Hieu Le, Srijan Das, Stephanie Schuckers \\}
    The University of North Carolina at Charlotte \\
    {\tt\small wbondura@charlotte.edu}
    \vspace{2em}
\end{center}
]

\setcounter{section}{0}
\renewcommand{\thesection}{\Alph{section}}

\section{Overview}

The Supplementary material is organized as follows:
\begin{itemize}
    \item Section \ref{sec:quant}: Additional Quantitative Results
    \item Section \ref{sec:qual}: Additional Qualitative Results
\end{itemize}

\section{Additional Quantitative Results}
\label{sec:quant}

\subsection{Ablation Studies}
Additional results for variants of our model, each trained for five epochs, are provided in Table \ref{tab:combined_ablation_celeb}. It is further evident that 4 DDIM steps provide the best balance during training and that both identity and landmark features are required for training a face swapping model of this nature.

% Single-column table locked to the text
\begin{table}[ht]
\centering
\caption{Ablation studies on 3D Condition Generation and DDIM Steps/Loss Components on CelebA dataset.}
\label{tab:combined_ablation_celeb}
\vspace{0.5em} % Adds clean spacing below the caption
\resizebox{\linewidth}{!}{
\begin{tabular}{lccccc}
\toprule
\multirow{2}{*}{\textbf{Method}} & \multirow{2}{*}{\textbf{FID}$\downarrow$} & \multicolumn{2}{c}{\textbf{ID retrieval} $\uparrow$} & \multirow{2}{*}{\textbf{Pose}$\downarrow$} & \multirow{2}{*}{\textbf{Expr.}$\downarrow$} \\
\cmidrule(lr){3-4}
& & \textbf{Top-1} & \textbf{Top-5} & & \\
\midrule
\multicolumn{6}{c}{\textit{3D Condition Generation}} \\
\midrule
\rowcolor{lightblue}\textbf{Avg. Pool} & \underline{5.78} & \textbf{79.30\%} & \textbf{90.40\%} & 3.73 & \underline{0.99} \\
Coarse & 6.46 & 44.30\% & 66.70\% & 3.88 & \textbf{0.97} \\
Middle & 7.24 & 54.40\% & 74.60\% & 3.87 & 1.05 \\
Fine-Grained & \textbf{5.53} & \underline{64.40\%} & \underline{82.70\%} & \textbf{3.56} & \underline{0.99} \\
Linear Proj. & 6.20 & 55.00\% & 75.60\% & \underline{3.71} & 1.01 \\
\midrule
\multicolumn{6}{c}{\textit{DDIM Steps and Loss Components}} \\
\midrule
\rowcolor{lightblue}\textbf{DiffSwap++} & \textbf{5.78} & \textbf{79.30\%} & \textbf{90.40\%} & \textbf{3.73} & \textbf{0.99} \\
DiffSwap++ (w/o ID Feat.) & 10.6 & 0.60\% & 0.20\% & 5.05 & 1.09 \\
DiffSwap++ (w/o LM Feat.) & 60.19 & 0.60\% & 1.20\% & 21.03 & 2.50 \\
DiffSwap++ (2 DDIM Steps) & 11.29 & 30.50\% & 53.10\% & 5.06 & 1.25 \\
DiffSwap++ (6 DDIM Steps) & 36.89 & 12.20\% & 33.50\% & 4.46 & 1.39 \\
\bottomrule
\end{tabular}
}
\end{table}

\subsection{Facial Recognition Evaluation Beyond ArcFace}
To further validate the robustness of our identity preservation, we expand our face recognition evaluation by incorporating the CosFace \cite{cosface} recognition model. As shown in Table \ref{tab:cosface_retrieval}, \modelname~achieves a 96.80\% Top-1 retrieval accuracy, providing additional evidence for the robustness of our identity preservation. Although DiffFace \cite{diffface} and REFace \cite{reface} achieve marginally higher retrieval accuracies under this specific metric, their generated face swaps exhibit noticeably lower visual realism. This is corroborated by the FID results in the main manuscript, where \modelname~achieves the lowest FID compared to REFace and DiffFace.

% Single-column table locked to the text
\begin{table}[ht]
\centering
\caption{Identity retrieval accuracy utilizing the CosFace recognition model.}
\label{tab:cosface_retrieval}
\vspace{0.5em} % Adds clean spacing below the caption
\resizebox{\linewidth}{!}{
\begin{tabular}{l ccccccc}
\toprule
\textbf{Metric} & \textbf{Ours} & \textbf{DiffFace} & \textbf{REFace} & \textbf{DiffSwap} & \textbf{HifiFace} & \textbf{SimSwap} & \textbf{FaceDancer} \\
\midrule
\textbf{Top-1 (\%)} & \textbf{96.80} & 98.20 & 98.02 & 61.80 & 95.50 & 93.90 & 88.20 \\
\textbf{Top-5 (\%)} & \textbf{99.10} & 99.40 & 99.67 & 79.81 & 97.80 & 94.80 & 96.30 \\
\bottomrule
\end{tabular}
}
\end{table}

\subsection{Dataset Illumination Variance}
Table \ref{tab:lighting_variance} displays extracted spherical harmonics lighting coefficients using SfSNet \cite{sfsnet}. We measured standard deviations across the X, Y, and Z axes, alongside overall directional intensity representing lighting contrast and harshness. We display that the primary datasets utilized in our methodology (CelebA \cite{celebA} and FFHQ \cite{ffhq}) exhibit significantly greater spatial variances than CFD \cite{cfd}, a controlled face dataset where each image has the same flat lighting. These metrics empirically confirm that our chosen in-the-wild test datasets successfully capture real-world illumination conditions.

% Single-column table locked to the text
\begin{table}[ht]
\centering
\caption{Illumination variance across datasets measured via spherical harmonics.}
\label{tab:lighting_variance}
\vspace{0.5em} % Adds clean spacing below the caption
\resizebox{\linewidth}{!}{
\begin{tabular}{l cccc}
\toprule
\textbf{Dataset} & \textbf{Lateral ($X$)} & \textbf{Vertical ($Y$)} & \textbf{Depth ($Z$)} & \textbf{Intensity ($\sigma$)} \\
\midrule
CFD & 0.0596 & 0.0342 & 0.0711 & 0.0353 \\
CelebA & 0.1044 & 0.1119 & 0.0856 & 0.0973 \\
FFHQ & 0.1245 & 0.1254 & 0.0893 & 0.1100 \\
\bottomrule
\end{tabular}
}
\end{table}

\section{Additional Qualitative Results}
\label{sec:qual}

\subsection{Extended Baseline Comparisons}
Qualitative comparisons for images generated from the AFLW2000-3D dataset \cite{AFLW}, which feature more extreme pose and expression, are provided in Figure \ref{fig:extreme}. We also provide extended comparisons from the CelebA \cite{celebA} dataset in Figure \ref{fig:celeb_comp}. These figures highlight the capability of \modelname~to produce face swaps which are consistently more realistic and better preserve the identity of the source face compared to other popular models.

Furthermore, Figure \ref{fig:simswap_supp} provides a visually prominent comparison specifically between \modelname~and SimSwap \cite{simswap} on the CelebV-Text \cite{celebvtext} dataset. As highlighted in the figure, SimSwap results often feature minimal changes to the target faces and introduce visual artifacts. This failure to properly integrate the source identity directly corresponds to their lower identity retrieval scores compared to \modelname.

% =================================================================
% INLINE SMALL FIGURES
% We manually set the counter to 2 so the next figure is Figure 3.
% =================================================================
\setcounter{figure}{2}

\begin{figure}[ht]
    \centering
    \includegraphics[width=\linewidth]{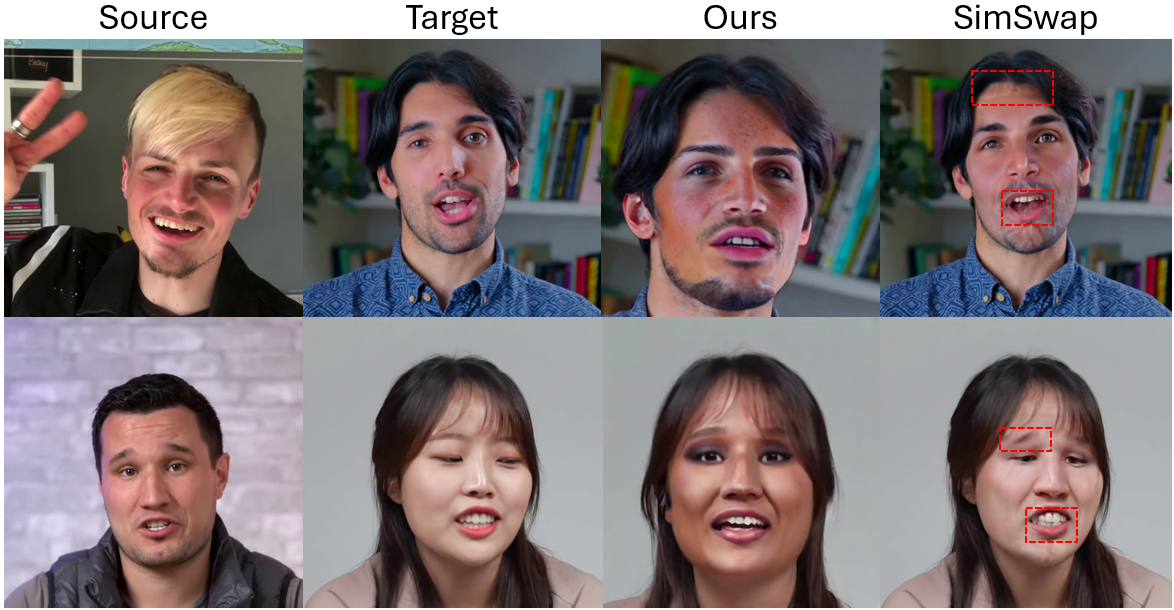}
    \caption{Targeted visual comparison highlighting artifact generation and lack of identity transfer in SimSwap compared to \modelname.}
    \label{fig:simswap_supp}
\end{figure}

\subsection{User Study Examples}
\begin{figure}[ht]
    \centering
    \includegraphics[width=\linewidth]{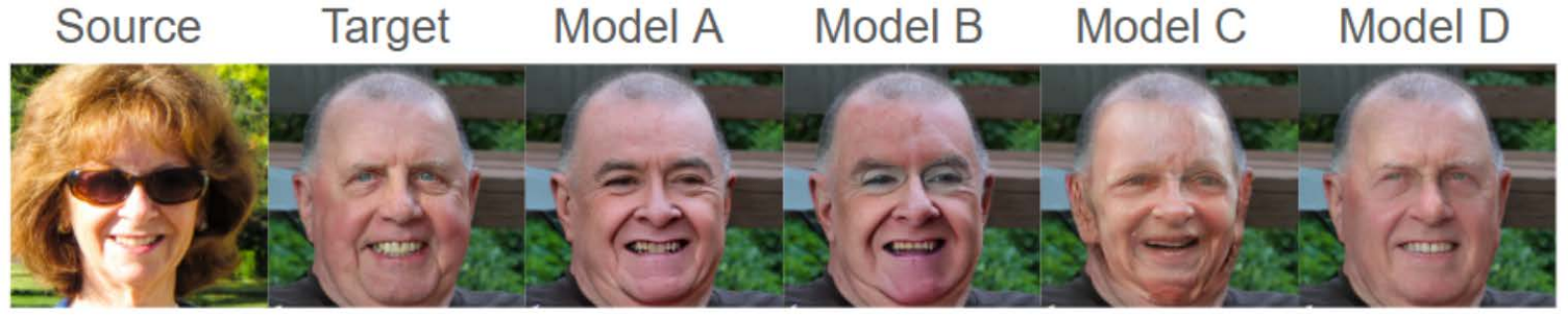}
    \caption{A representative example of the presentation format utilized during the human evaluation user study.}
    \label{fig:user_study_supp}
\end{figure}

Figure \ref{fig:user_study_supp} provides a representative visual sample of the interface and images used in our user study. To ensure a fair and consistent evaluation, we confirm that all images presented to participants were of identical quality, including the exact same resolution, aspect ratio, and display settings. 

% =================================================================
% FLUSH TEXT AND INLINE ELEMENTS
% This completely clears the queue before we define the large figures
% =================================================================
\clearpage

% =================================================================
% LARGE FULL-PAGE FIGURES
% We manually set the counter to 0 so the next figure is Figure 1.
% =================================================================
\setcounter{figure}{0}

\begin{figure*}[p]
    \centering
    \includegraphics[width=\linewidth]{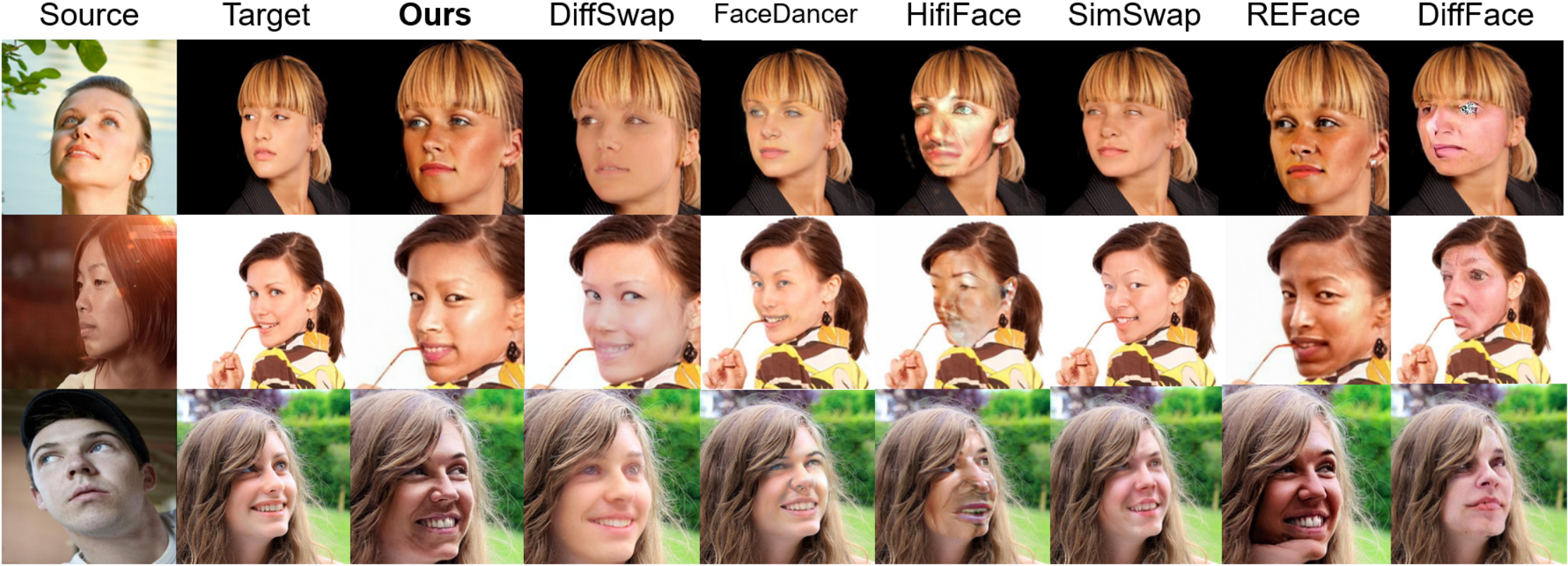}
    \caption{Qualitative comparison for extreme pose and expression.}
    \label{fig:extreme}
\end{figure*} 

\begin{figure*}[p]
    \centering
    \includegraphics[width=\linewidth]{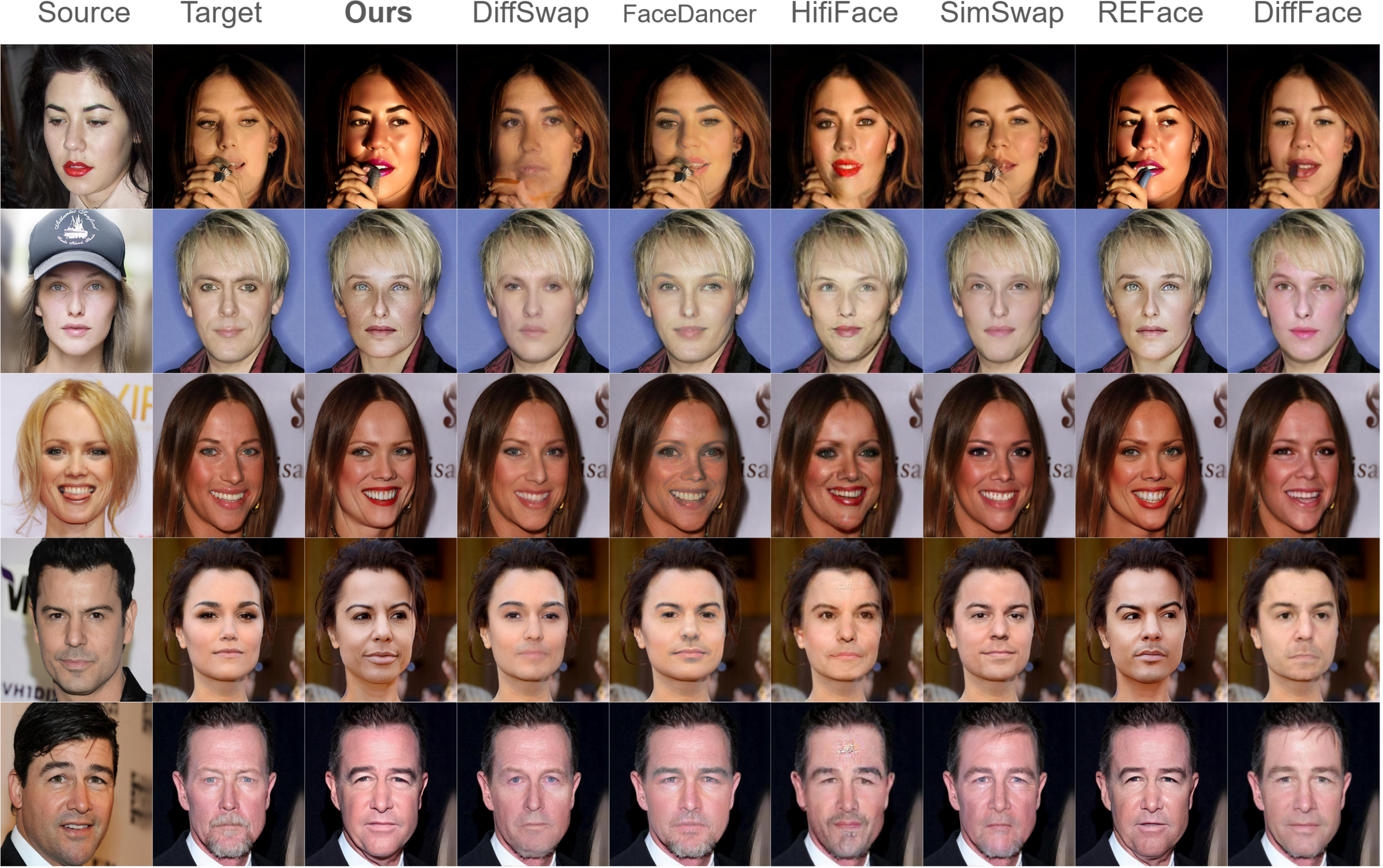}
    \caption{Qualitative comparison of all models on CelebA.}
    \label{fig:celeb_comp}
\end{figure*} 

% =================================================================
% FLUSH LARGE FIGURES
% Ensures Figures 1 & 2 print on their own pages BEFORE references
% =================================================================
\clearpage

\end{document}